\PassOptionsToPackage{table,xcdraw}{xcolor}

\documentclass[lettersize,journal]{IEEEtran}
\usepackage{hyperref}       
\usepackage{amsmath}
\usepackage{amsfonts}
\usepackage{amssymb}
\usepackage{algorithm}
\usepackage{algpseudocode}
\usepackage{xspace}
\usepackage{bm}
\usepackage[textsize=footnotesize]{todonotes}
\usepackage{color}

\usepackage{comment}

\usepackage{booktabs}
\usepackage{acronym}
\usepackage{cleveref}

\usepackage{multirow}

\usepackage{booktabs}
\usepackage[table,xcdraw]{xcolor}

\usepackage{subcaption} 
\usepackage{caption}    
\newcommand{\myparagraph}[1]{~\\ \noindent \textbf{#1}.}
\newcommand{\multishield}{\emph{Multi-Shield}\xspace}

\newcommand{\rc}{\rowcolor[HTML]{EFEFEF}}

\newcommand{\standardshort}{\texttt{C1}\xspace} 
\newcommand{\carmonshort}{\texttt{C2}\xspace} 
\newcommand{\chenshort}{\texttt{C3}\xspace} 
\newcommand{\gowalshort}{\texttt{C4}\xspace} 
\newcommand{\addepallishort}{\texttt{C5}\xspace} 
\newcommand{\xushort}{\texttt{C6}\xspace} 
\newcommand{\debenedettishort}{\texttt{I1}\xspace} 
\newcommand{\liushort}{\texttt{I2}\xspace} 
\newcommand{\singhconvnextshort}{\texttt{I3}\xspace} 
\newcommand{\singhvitbshort}{\texttt{I4}\xspace} 

\newcommand{\resnet}{\texttt{M1}\xspace}
\newcommand{\wideres}{\texttt{M2}\xspace}
\newcommand{\vit}{\texttt{M3}\xspace}

\newcommand{\standardshortcit}{\texttt{C1}~\citep{Croce2021Robustbench}\xspace} 
\newcommand{\carmonshortcit}{\texttt{C2}~\citep{Carmon-19-unlabeled}\xspace} 
\newcommand{\chenshortcit}{\texttt{C3}~\citep{Chen2020Robust}\xspace} 
\newcommand{\gowalshortcit}{\texttt{C4}~\citep{Gowal2021Robust}\xspace} 
\newcommand{\addepallishortcit}{\texttt{C5}~\citep{Addepalli22Robust}\xspace} 
\newcommand{\xushortcit}{\texttt{C6}~\citep{Xu2023Robust}\xspace} 
\newcommand{\debenedettishortcit}{\texttt{I1}~\citep{debenedetti2023Robust}\xspace} 
\newcommand{\liushortcit}{\texttt{I2}~\citep{liu2024comprehensive}\xspace} 
\newcommand{\singhconvnextshortcit}{\texttt{I3}~\citep{singh2024revisiting}\xspace} 
\newcommand{\singhvitbshortcit}{\texttt{I4}~\citep{singh2024revisiting}\xspace} 

\def\equationautorefname~#1\null{Eq.~(#1)\null}

\usepackage[square,comma,sort]{natbib}

\begin{document}
%
\title{Robust image classification with multi-modal large language models}
%
%
%

\DeclareRobustCommand{\IEEEauthorrefmark}[1]{\smash{\textsuperscript{\footnotesize #1}}}
\author{
\IEEEauthorblockN{
Francesco Villani\IEEEauthorrefmark{a},
Igor Maljkovic$^*$\IEEEauthorrefmark{a},
Dario Lazzaro$^*$\IEEEauthorrefmark{b}\IEEEauthorrefmark{a},
Angelo Sotgiu\IEEEauthorrefmark{c},
Antonio Emanuele Cinà$^\dagger$\IEEEauthorrefmark{a},
Fabio Roli\IEEEauthorrefmark{a}}\\
\vspace{0.05in}
\IEEEauthorblockA{\IEEEauthorrefmark{a} University of Genoa, Via Dodecaneso 35, Genoa, 16145, Italy}\\
\IEEEauthorblockA{\IEEEauthorrefmark{b} Sapienza University of Rome, Via Ariosto 25, Rome, 00185, Italy}\\
\IEEEauthorblockA{\IEEEauthorrefmark{c} University of Cagliari, Via Marengo 3, Cagliari, 09100, Italy}
\vspace{0.05in}

\thanks{$^\dagger$ Corresponding author:
antonio.cina@unige.it (Antonio Emanuele Cinà)}
\thanks{$^*$Authors contributed equally:}
\thanks{e-mail: francesco.villani@edu.unige.it (Francesco Villani),
igor.maljkovic@edu.unige.it (Igor Maljkovic), dario.lazzaro@uniroma1.it
(Dario Lazzaro), angelo.sotgiu@unica.it (Angelo Sotgiu), fabio.roli@unige.it (Fabio Roli)}
}

\maketitle

\begin{abstract}
Deep Neural Networks are vulnerable to adversarial examples, i.e., carefully crafted input samples that can cause models to make incorrect predictions with high confidence. 
To mitigate these vulnerabilities, adversarial training and detection-based defenses have been proposed to strengthen models in advance. However, most of these approaches focus on a single data modality, overlooking the relationships between visual patterns and textual descriptions of the input.
In this paper, we propose a novel defense, \multishield, designed to combine and complement these defenses with multi-modal information to further enhance their robustness. \multishield leverages multi-modal large language models to detect adversarial examples and abstain from uncertain classifications when there is no alignment between textual and visual representations of the input. Extensive evaluations on CIFAR-10 and ImageNet datasets, using robust and non-robust image classification models, demonstrate that \multishield can be easily integrated to detect and reject adversarial examples, outperforming the original defenses.
\end{abstract}


\IEEEpeerreviewmaketitle

\section{Introduction} \label{sec:introduction}
\IEEEPARstart{D}{eep} Neural Networks (DNNs) are employed in numerous image recognition tasks~\citep{rawat2017deep}, such as facial recognition, medical imaging, pedestrian segmentation for autonomous driving, and surveillance. 
For instance, in medical imaging, DNNs assist in detecting diseases like cancer from X-rays and MRIs~\citep{mall2023comprehensive}. In autonomous driving~\citep{yurtsever2020survey}, DNNs help recognize pedestrians and road signs, improving safety and navigation.
Their remarkable performance has, therefore, led to widespread integration into various applications, impacting user safety, daily routines, services, and product quality. 
However, these models are vulnerable to adversarial examples~\citep{biggio2013evasion, szegedy2013intriguing, carlini2017towards}—subtle, often imperceptible perturbations in the input data that can mislead the model into making incorrect predictions or classifications. 
Adversarial examples are typically generated through gradient-based evasion attacks~\cite{croce2020reliable,cina2024sigma,cina2024attackbench}, which exploit the model's sensitivity to minor input changes to induce erroneous predictions, often with high confidence.
Furthermore, adversarial examples pose a significant threat to the reliability of DNNs, especially in user-facing and safety-critical applications. 
These attacks not only degrade model performance but also introduce serious concerns about compliance with emerging regulatory standards. 
For instance, frameworks like the AI Act~\citep{EU2019Trustworthy} have imposed stricter demands on the reliability and accountability of AI systems. 
Therefore, as DNNs deployment expands and regulatory requirements tighten, developing robust defenses against adversarial examples is increasingly important for safe and responsible use.

Consequently, various defensive strategies have been proposed in recent years to address these challenges.
Prominent among them are adversarial training and detectors, complementary approaches both representing proactive defenses that align with the \emph{security-by-design} principle by incorporating security measures during model development~\citep{biggio2018wild}. 
Adversarial training~\citep{goodfellow2014explaining,madry2017towards} strengthens models by injecting knowledge of adversarial attacks during their training (e.g., including adversarial examples in the training data), improving their resilience to future attacks.
Detector-based defenses, on the other hand, encompass techniques that detect and mitigate/reject attacks as they occur. Examples include anomaly detection systems that reject potentially dangerous inputs in real-time and abstain from predicting them~\citep{lu2017safetynet,melis2017deep,sotgiu2020deep}.
However, each of these defense strategies has its own limitations. Adversarial training, while effective, can be complex to develop and resource-intensive~\citep{shafahi2019adversarial}, requiring continuous adaptation or retraining to address evolving attacks. Conversely, detectors rely on recognizing deviations from normal behavior, which may not be effective against new attacks~\citep{biggio2018wild}.

To address these challenges, we propose \multishield, a solution that integrates the strengths of both adversarial training and detection defenses while handling multimodal information.  The idea is to combine adversarial training defenses with a rejection module that incorporates both visual and textual information during prediction. 
In particular, \multishield uses an adversarially trained model and a multimodal model (i.e., CLIP~\citep{radford2021learning}), which analyzes visual and textual patterns to verify their alignment. Specifically, the multimodal model uses visual images and textual descriptions related to the classes in the dataset to assess semantic agreement, employing zero-shot predictions.
As illustrated in \autoref{fig:multishield}, when there is an agreement between the proactively trained image classifier and the multimodal model, the system provides a prediction. 
However, if there is a discrepancy, the system abstains from responding, as the lack of agreement suggests potential adversarial manipulation or uncertainty. 
Our experiments demonstrate the effectiveness of \multishield on six benchmark datasets (i.e., CIFAR-10, ImageNet, Caltech-101, Food-101, Oxford-IIIT pets, STL-10). We applied \multishield to both non-robust models and robust models developed with adversarial training.
The results reveal that \multishield significantly increases robustness for proactively trained models, as it successfully rejects adversarial examples that would otherwise bypass them. Overall, these results highlight \multishield's ability to complement existing defenses and provide an additional layer of security for them.

In this work, we make the following key contributions: 
\begin{itemize}
    \item we propose \multishield, a novel defense mechanism that leverages multimodal data, specifically visual and textual inputs, to identify and reject adversarial examples;

    \item we explore the interaction between adversarial training and a multimodal detector, demonstrating how \multishield can be effectively integrated as a supplementary layer to enhance overall model robustness;

    \item we assess the robustness improvements provided by multimodal information against adversarial attacks by evaluating \multishield on six datasets and thirteen models.
\end{itemize}

The rest of the paper is structured as follows: \autoref{sec:background} provides an overview of adversarial defenses and the principles of multimodal large language models. In \autoref{sec:methodology}, we present the architecture and inner workings of \multishield. \autoref{sec:experiments} outlines our experimental setup and results, and \autoref{sec:conclusions} concludes with limitations and future research directions.

\section{Background and Related Work}\label{sec:background}
In this section, we present background information and review related work on adversarial attacks and defenses, as well as multimodal large language models.
\subsection{Adversarial examples}
Adversarial examples are carefully crafted inputs designed to mislead models into making incorrect predictions, with the purpose of revealing and exploiting vulnerabilities in machine learning models. 
Given an input sample $\bm{x} \in \mathcal{X} = [0,1]^d$ with its true label $y \in \mathcal{Y} = \{1, \ldots, K\}$ (where $K$ is the number of classes) and a trained classifier $f: \mathcal{X} \to \mathcal{Y}$, an adversarial example $\bm{x}'$ is a slightly perturbed version of $\bm{x}$, designed to cause misclassification. Formally, the adversarial example can be found by solving the following constrained optimization problem:
\begin{align}
    \label{eq:min_prob}\min_{\bm{x}'} & \; L(\bm{x}', y) \\ 
    \label{eq:lp_projection}\text{s.t.} & \; \|\bm{x}' - \bm{x}\|_{p} \leq \varepsilon \\
    \label{eq:box_constraint}& \; \bm{x}' \in [0,1]^d ,\
\end{align}
where $L(\cdot)$ in \autoref{eq:min_prob} is a loss function that penalizes correct classification of $\bm{x}'$ by $f$. A commonly used loss function is the Difference of Logits~\citep{carlini2017towards}, expressed as:
\begin{align} \label{eq:dl_loss}
    L(\bm{x}, y) = f_{y}(\bm{x}) - \max_{j \in \mathcal{Y}\setminus \{y\}} f_{j}(\bm{x}),
\end{align}
with $ f_i(\bm{x}) $ denoting the logit score of $\bm{x}$ associated with class $i$.
Under this loss, the optimization program aims to find an adversarial example $\bm{x}'$ that decreases the logit corresponding to the true class label $y$ while increasing the logits of competing incorrect classes. As a result, the model is driven to misclassify the input by predicting the incorrect class with the highest logit score.
The constraint in  \autoref{eq:lp_projection}  ensures that the adversarial example $\bm{x}'$ remains similar to the original input $\bm{x}$ by limiting the norm of their difference, with typical choices for $p \in \{0, 1, 2, \infty\}$~\citep{carlini2017towards}. 
Lastly, the constraint in \autoref{eq:box_constraint} ensures that $\bm{x}'$ remains within the valid input space $[0,1]^d$, a typical requirement for image recognition tasks.
Given the above formulation, several optimization methods have been proposed to solve this problem, including PGD~\citep{MadryMSTV18}, FGSM~\citep{goodfellow2014explaining}, and many others~\citep{cina2024attackbench}. 
Among them, AutoAttack~\citep{croce2020reliable} stands out as a state-of-the-art approach due to its ability to generate adversarial examples effectively without parameter tuning, making it well-suited for reliable and consistent benchmarking of model robustness.

\subsection{Adversarial Defenses}
Prominent proactive defenses against adversarial examples in DNNs fall into two main categories: robust optimization techniques and adversarial detectors~\citep{biggio2018wild,sotgiu2020deep}.
Among the most effective methods in the first category is adversarial training~\citep{goodfellow2014explaining, madry2017towards,li2024adversarial}, which strengthens robustness by exposing the model to adversarially perturbed inputs during the training process and has consistently proven to be one of the most effective defense mechanisms to date~\citep{madry2017towards,croce2020reliable}.
Detectors, on the other hand, focus on mechanisms that enable the model to abstain from making predictions when it suspects inputs to be adversarial. The pioneering approach by \cite{bendale2016towards} involves rejecting samples whose feature representations are significantly distant from class centroids. Similarly, \cite{melis2017deep} propose a distance-based rejection method using thresholds applied to outputs of a multi-class SVM 
to estimate data feature representation distribution.
Later works extend the inspection of feature representations beyond the final layer of the network. \cite{lu2017safetynet} leverage late-stage ReLU activations combined with a quantized RBF-SVM to detect adversarial examples, though this method requires adversarial training, increasing computational cost. 
Alternatively, \cite{papernot2018deep} present a KNN-based method that evaluates intermediate layer representations and rejects inputs with inconsistencies, though it incurs substantial overhead due to the need for distance computations with respect to training data at each layer. \cite{sotgiu2020deep} propose a method for evaluating the consistency of feature representations across different layers, using a threshold for distance-based rejection.

However, all these strategies are restricted to single-modality data. They focus solely on the visual semantics of the input during training, overlooking the semantical alignment between predicted classes and visual patterns.
Furthermore, the interaction between adversarial training and rejection remains unexplored, preventing the development of a more robust defense.

\subsection{Multimodal Vision-Language Models}
Multimodal models represent a recent development in AI, enabling the simultaneous processing of multiple types of data~\citep{yin2023survey}. These models are trained to process data from different modalities, identifying patterns aligning the information. 
In particular, vision-language map images and textual descriptions into a shared semantic space, facilitating tasks that require a joint understanding of both modalities~\citep{du2022survey}. Applications of these models include image captioning~\citep{hossain2019comprehensive}, visual question answering~\citep{wu2017visual}, and text-to-image retrieval~\citep{cao2022image}, where the combination of visual and textual information allows models to better capture contextual meaning and improve performance.
In this work, we focus on vision-language models applied to image recognition tasks.
Several prominent vision-text multimodal models have emerged, including VisualBERT~\citep{li2019visualbert}, VilBERT~\citep{lu2019vilbert}, and VisionLLaMA~\citep{chu2024visionllama}, each designed to capture complex relationships between visual and textual inputs. These models typically employ transformer-based architectures~\citep{vaswani2017attention} to encode visual and textual data into a unified representation. 
Another notable example of vision-text multimodal is CLIP (Contrastive Language-Image Pretraining)~\citep{radford2021learning}, which aligns visual and textual embeddings using large-scale datasets of paired images and captions. 
CLIP works by training on a contrastive learning objective, ensuring that semantically similar images and text are positioned closely within a shared embedding space, leading to strong performance in tasks such as zero-shot image classification and text-to-image retrieval~\citep{radford2021learning}.

\begin{figure*}[h]
    \centering
    \includegraphics[width=0.48\textwidth]{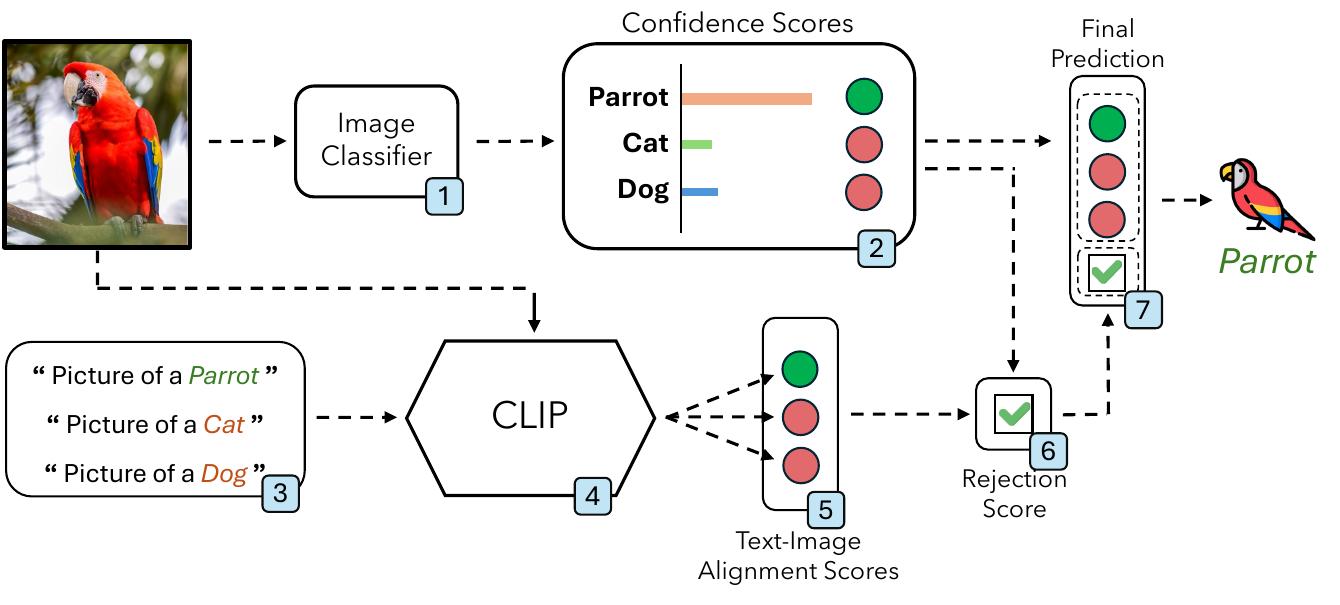}
    \hfill
    \vline
    \hfill
    \includegraphics[width=0.48\textwidth]{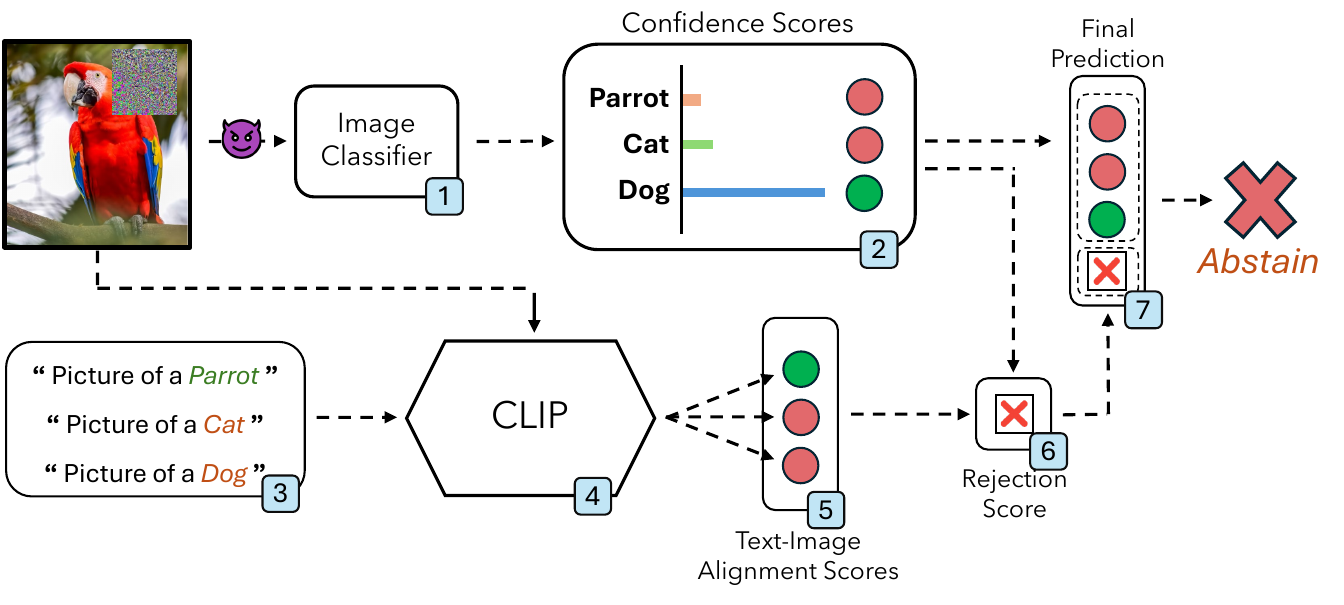}
\caption{Illustration of \multishield's operation. An image classifier (1) first processes the input image to generate an initial prediction (2). Simultaneously, the CLIP model (4) performs zero-shot classification using text prompts (3). A rejection score (6) is computed by comparing the agreement between the image classifier's prediction (2) and the CLIP model's output (5). Lastly, \multishield either returns the final prediction or abstains (7), according to the uncertainty in classification. The left part illustrates a clean prediction, while the right part shows \multishield abstaining for an adversarial example.}
    \label{fig:multishield}
\end{figure*}

\section{\multishield's Rejection Mechanism}\label{sec:methodology}
We now present the \multishield structure, detailing its operations, problem formulation, and the design of an adaptive attack to evaluate its robustness under worst-case scenarios.

\subsection{Multishield's Workflow}
The \multishield framework operates in three distinct phases, depicted in \autoref{fig:multishield}, each contributing to a robust image classification process that leverages both unimodal and multimodal analysis. 
We name these phases with: \emph{unimodal prediction}, \emph{multimodal alignment}, and \emph{Multi-Shield decision}.\medskip
\myparagraph{Unimodal Prediction}
The \multishield workflow begins with the image classifier processing the input (Step 1) and generating confidence scores for each class in the dataset (Step 2). 
Based on these scores, an initial prediction is made, relying solely on the image classifier's output. 
Notably, the image classifier at this stage can be either a standard (non-robust) model or a robust model trained to resist adversarial attacks. 
In either case, \multishield operates on top of the classifier, augmenting its defense with an additional verification layer.
\medskip
\myparagraph{Multimodal Alignment}
In this second phase, \multishield uses the CLIP model as a zero-shot vision-language classifier~\citep{radford2021learning}. It compares the visual representation of the input image with class description prompts to determine alignment. 
For each class, we automatically create natural language prompts in the format \texttt{`Picture of a [object]'} (Step 3). For example, for classes such as `parrot' and `dog,' the prompts would be \texttt{`Picture of a parrot'} and \texttt{`Picture of a dog'}, respectively.
The text prompts serve as a simplified but effective method to represent each class conceptually; however, with some human effort, they could be further enriched by including additional descriptive characteristics (e.g., mentioning that parrots have feathers and wings).
During prediction, these prompts, along with the input image, are processed through CLIP’s dual encoders: one for images and one for text (Step 4). This process generates visual and textual embeddings that capture the features of both modalities, with CLIP aiming to maximize the alignment between the image and the textual description that best matches it.
Finally, \multishield performs zero-shot classification with CLIP by computing alignment scores using cosine similarity and selecting the class with the highest score as the final prediction~\citep{radford2021learning} (Step 5).\medskip
\myparagraph{Multi-Shield Decision}
In the final phase, \multishield compares the predictions from the unimodal image classifier and the multimodal CLIP model (Step 6), each capturing distinct patterns in the data. If both models agree, \multishield outputs their shared prediction; if not, it abstains (Step 7). 
As shown on the left side of \autoref{fig:multishield}, for a clean image, both models confidently identify the class `parrot,' leading \multishield to output the agreed-upon prediction.
Conversely, on the right side of \autoref{fig:multishield}, adversarial noise causes the unimodal classifier to misclassify the image, while the CLIP model still correctly aligns the image with the `parrot' class. This disagreement drives \multishield to abstain, acknowledging that the adversarial input may have misled one model.
Unlike typical robust models, which do not employ a rejection mechanism, \multishield introduces this feature to abstain when uncertain (e.g., input is suspected to be adversarial) and ensures predictions are made only when both models agree.

\subsection{Problem Formulation} 
\multishield is defined as a classification function $\hat{f}: \mathcal{X} \to \hat{\mathcal{Y}}$, where $\hat{\mathcal{Y}} = \mathcal{Y} \cup \{K+1\}$ denotes the set of class labels in the dataset $\mathcal{Y}$  extended with an additional \emph{rejection class}. 
At inference time, for each input sample $\bm{x} \in \mathcal{X}$, \multishield computes a rejection score $R(\bm{x})$, which determines whether to make a prediction or abstain. The final decision of $\hat{f}$ is then given by: 
\begin{align} 
\hat{f}(\bm{x}) = 
\begin{cases} 
f(\bm{x}), & \text{if } R(\bm{x}) \leq 0 \\ 
k+1, & \text{if } R(\bm{x}) > 0 .\ 
\end{cases} 
\end{align}
Here, $R(\bm{x})$ represents the level of agreement between the unimodal image classifier and the zero-shot multimodal prediction from CLIP, serving as a score to gauge the model's uncertainty about the input. 
When $R(\bm{x})$  is non-positive, it indicates that the image classifier and CLIP are in agreement, suggesting high confidence in the prediction. Conversely, if $R(\bm{x})$  is positive, the model abstains from making a prediction due to uncertainty.
The rejection score $R(\bm{x})$ of \multishield is computed as:
\begin{align} \label{eq:rejection_threshold}
R(\bm{x}) = \left| \max_{i \in \{1,\ldots,K\}} h(\bm{x}, P_i) - h(\bm{x}, P_j) \right|,
\end{align} where $j = f(\bm{x})$ is the predicted class from the image classifier and $P_i$ represents the textual prompt for class $i$. The function $h$ measures the text-image alignment scores using CLIP's zero-shot capabilities, computing the cosine similarity between the embeddings of the image and the textual prompts in a shared semantic space. Formally: 
\begin{align}
h(\bm{x}, P_i) = \frac{h_{\rm{img}}(\bm{x})\cdot h_{\rm{txt}}(P_i)}{\lvert h_{\rm{img}}(\bm{x}) \rvert \lvert h_{\rm{txt}}(P_i)\rvert} 
\end{align} 
where $h_{\rm{img}}(\bm{x}) \in \mathbb{R}^d$ is the visual embedding of the input image and $h_{\rm{txt}}(P_i) \in \mathbb{R}^d$ is the textual embedding for the class prompt $i$, both generated using the encoders of the multimodal model. 
\multishield evaluates the alignment score for each class prompt and picks the highest one to perform zero-shot classification using the CLIP model. 
Finally, when \multishield abstains, the positive score in $R(\bm{x})$ indicates the level of disagreement between the image classifier and CLIP’s classification, supporting decisions on abstention reliability and subsequent input handling.

\subsection{Attacking \multishield}
To evaluate the worst-case robustness of \multishield, we assess its performance against an adaptive attacker~\citep{carlini2017adversarial}, who is fully aware of the defense and actively seeks to circumvent it. Adaptive attacks indeed provide a more rigorous and stronger test of defenses~\citep{papernot2018deep, carlini2019evaluating,sotgiu2020deep}.
The attacker, having access to the image recognition model and the rejection mechanism in \multishield, aims to simultaneously deceive the unimodal image classifier and manipulate the zero-shot multimodal classifier, causing both to agree on an incorrect label.
The adaptive attack can thus be formalized by revisiting the optimization program in \autoref{eq:min_prob}-\autoref{eq:box_constraint} as follows:
\begin{align}
    \min_{\bm{x}'} & \; L(\bm{x}', y) \\ 
    \text{s.t.} & \; \|\bm{x}' - \bm{x}\|_{p} \leq \varepsilon \\
    & \; \bm{x}' \in [0,1]^d \\
    & \label{eq:adaptive_constraint} R(\bm{x}') \leq 0 \, 
\end{align}
The additional constraint in \autoref{eq:adaptive_constraint} ensures that the adversarial input does not trigger the rejection mechanism of \multishield.

\section{Experiments}\label{sec:experiments} In this section, we evaluate the effectiveness of \multishield's rejection mechanism in defending against adversarial attacks. Our goal is to assess \multishield's performance under both traditional (non-adaptive) and adaptive attack scenarios.
\begin{table*}[t]
\centering
\caption{Experimental results on baseline and adversarially pre-trained models (6 on CIFAR-10, 4 on ImageNet, and 3 on Caltech-101, Food-101, Oxford-IIIT Pets, and STL-10) evaluated using clean accuracy (\%), robust accuracy (\%), and rejection ratio (\%). The defense is assessed using AutoAttack with perturbation size $\varepsilon = 8/255$ across three defense scenarios: without \multishield (\textbf{Baseline}), with \multishield (\textbf{Multi-Shield}), and adaptively attacking \multishield (\textbf{Multi-Shield - Adaptive Attack}). Note that performing the attack on either of the underlying clip models yields a robust accuracy of 0\%. \smallskip}
\label{tab:multishield_effectiveness}
\setlength\tabcolsep{7.pt}
\renewcommand{\arraystretch}{.95}
\newcolumntype{L}{>{\hspace*{-\tabcolsep}}c}
\newcolumntype{R}{c<{\hspace*{-\tabcolsep}}}
\begin{tabular}{L@{}cccccccc@{}R}
\toprule
 & \multicolumn{2}{c}{\textbf{Baseline}} & \multicolumn{3}{c}{\textbf{Multi-Shield}} & \multicolumn{2}{c}{\textbf{Multi-Shield - Adaptive Attack}} \\ \cmidrule(l){2-3} \cmidrule(l){4-6} \cmidrule(l){7-8}
Models & \xspace\xspace Clean Acc. & Robust Acc. & Clean Acc. & Robust Acc. & Rejection Ratio  & Robust Acc.  & Rejection Ratio  \\ \midrule
 \multicolumn{8}{c}{\textbf{CIFAR-10}} \\
\midrule
\rc
\standardshort & 94.8 & 0.0 & 92.7 & 91.8 & 91.8 & 20.4 & 0.4 \\
\carmonshort & 89.7 & 59.9 & 87.8 & 92.0 & 32.5 & 63.6 & 0.7 \\
\rc
\chenshort & 86.1 & 52.0 & 84.5 & 91.9 & 40.3 & 55.0 & 0.4 \\
\gowalshort & 88.7 & 66.8 & 87.0 & 93.3 & 27.2 & 69.6 & 0.8 \\
\rc
\addepallishort & 85.7 & 52.9 & 84.2 & 92.2 & 39.5 & 58.4 & 0.5 \\
\xushort & 93.7 & 65.6 & 91.6 & 88.4 & 23.4 & 66.9 & 0.6 \\
\midrule
 \multicolumn{8}{c}{\textbf{ImageNet}} \\
\midrule
\rc
\debenedettishort & 70.9 & 10.3 & 63.8 & 88.7 & 78.7 & 17.4 & 4.4 \\
\liushort & 75.1 & 27.0 & 67.1 & 89.4 & 63.5 & 32.5 & 2.8 \\
\rc
\singhconvnextshort & 74.9 & 26.2 & 67.3 & 87.8 & 62.6 & 30.7 & 3.0 \\
\singhvitbshort & 74.7 & 22.7 & 66.7 & 87.9 & 66.0 & 28.4 & 2.7 \\

\midrule
\multicolumn{8}{c}{\textbf{Caltech-101}} \\
\midrule
\rc\resnet& 97.2 & 0.3 & 90.5 & 97.1 & 96.8 & 71.3 & 4.2 \\
\wideres & 97.4 & 0.2 & 90.9 & 97.7 & 97.6 & 67.3 & 3.3 \\
\rc\vit & 96.5 & 0.0 & 90.1 & 94.8 & 94.8 & 39.2 & 4.6 \\
\midrule
\multicolumn{8}{c}{\textbf{Food-101}} \\
\midrule
\rc\resnet& 80.3 & 0.0 & 77.3 & 97.7 & 97.7 & 74.4 & 26.2 \\
\wideres & 82.6 & 0.0 & 79.4 & 96.9 & 96.9 & 65.4 & 21.1 \\
\rc\vit & 86.2 & 0.0 & 83.4 & 98.6 & 98.6 & 27.9 & 12.4 \\
\midrule
\multicolumn{8}{c}{\textbf{Oxford-IIIT Pets}} \\
\midrule
\rc\resnet& 93.1 & 1.1 & 89.6 & 94.8 & 93.8 & 28.4 & 6.8 \\
\wideres & 93.3 & 1.1 & 89.8 & 94.5 & 93.4 & 22.2 & 6.1 \\
\rc\vit & 91.3 & 0.5 & 88.2 & 97.8 & 97.2 & 18.4 & 6.5 \\
\midrule
\multicolumn{8}{c}{\textbf{STL-10}} \\
\midrule
\rc\resnet& 97.9 & 0.3 & 97.2 & 98.8 & 98.5 & 61.2 & 0.5 \\
\wideres & 98.2 & 0.4 & 97.6 & 98.4 & 97.9 & 51.5 & 0.3 \\
\rc\vit & 97.8 & 0.0 & 97.0 & 99.2 & 99.2 & 17.2 & 0.1 \\
 \bottomrule
\end{tabular}
\end{table*}
\subsection{Experimental Setup} 
\myparagraph{Datasets}
We consider six datasets: CIFAR-10~\citep{Krizhevsky2009LearningML}, ImageNet~\citep{Krizhevsky2012ImageNetCW}, Caltech-101~\citep{fei2004learning}, Food-101~\citep{bossard2014food}, Oxford-IIIT Pets~\cite{parkhi2012cats}, STL-10~\cite{coates2011analysis}.
For CIFAR-10, we use all 10,000 test images. For ImageNet, we randomly sample $5000$ validation images. We evaluate the full validation sets for Caltech-101, Oxford-IIIT Pets, and STL-10, and sample $10000$ images for Food-101.
\medskip
\myparagraph{Classifiers}
For the image classifier, we source baseline and top state-of-the-art $\ell_{\infty}$ robust models from the RobustBench library~\citep{Croce2021Robustbench}. For CIFAR-10, we evaluate six models denoted with \standardshort-\xushort.  \standardshortcit is a non-robust model. \carmonshortcit,\gowalshortcit and \addepallishortcit employ adversarial training combined with data augmentation. \chenshortcit utilizes a robust ensemble of models, while \xushortcit reformulates adversarial training to maximize the margin for more vulnerable samples, enhancing overall robustness.
For ImageNet, we select four robust models, denoted with \debenedettishort -\singhvitbshort, incorporating adversarial training. \debenedettishortcit and \singhvitbshortcit utilize a ViT architecture, \liushortcit employs a Swin Transformer, and \singhconvnextshortcit uses the ConvNeXt architecture.
For experiments on Caltech-101, Oxford-IIIT Pets, STL-10, and Food-101, we fine-tune three ImageNet-pretrained models—ResNet-50 (\resnet), WideResNet50-2 (\wideres), and ViT-B/16 (\vit).
For each model, we replace the final classification layer and fine-tune the entire model for each dataset over $50$ epochs using stochastic gradient descent with a learning rate of 0.001 and momentum of $0.9$.
\medskip
\myparagraph{Multishield Construction}
The \multishield module combines a CLIP model with a standard image classifier. For CIFAR-10, we use a ViT-B visual encoder fine-tuned for CIFAR-10 classification~\citep{tang2024fusionbench}, and for ImageNet, a pre-trained ViT-L~\citep{li2023clipav2}, both paired with their respective text encoders from Hugging Face~\cite{HFtransformers}.\medskip
\myparagraph{Attack}
We conduct the experiments using AutoAttack~\citep{croce2020reliable}, a state of the art algorithm.  We use both a naive (non-adaptive) version of AutoAttack and an adaptive variant, which takes into account \multishield’s rejection mechanism.
Lastly, we set AutoAttack's perturbation size to $\varepsilon = 8/255$, as typical reference for robustness evaluation~\citep{Croce2021Robustbench}.\medskip
\myparagraph{Evaluation Metrics}
In our experiments, we assess \multishield's performance using several key metrics. Clean Accuracy measures the accuracy on unperturbed inputs, providing a baseline for the model's performance on pristine data. In this case, rejected samples are considered errors. The difference between the Clean Accuracy without any defense in place and \multishield's Clean Accuracy corresponds to the fraction of wrongly rejected samples (i.e., false positives). Robust Accuracy evaluates how well \multishield performs on adversarial inputs by accounting for correct predictions and cases where it abstains from making a prediction. In fact, the desired model behavior in the presence of adversarial examples is to avoid misclassification. The Rejection Ratio represents the percentage of inputs for which \multishield opts to abstain, reflecting the effectiveness of the rejection mechanism in identifying adversarial examples. The difference between the Robust Accuracy and Rejection Ratio corresponds to the fraction of correctly classified test inputs. 
Lastly, all experiments have been run on a NVIDIA A100 Tensor Core GPU with 40 GB of VRAM. 
\subsection{Experimental Results}
\begin{figure}[t]
    \centering
    \includegraphics[width=0.5\textwidth]{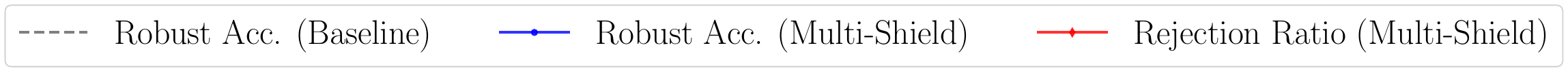}  
    \begin{subfigure}[b]{0.235\textwidth}
        \centering
        \includegraphics[width=\textwidth]{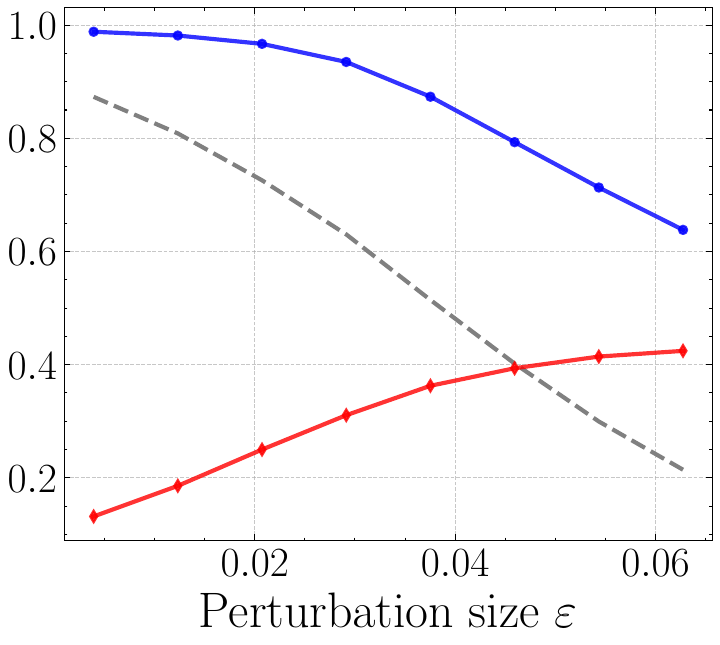}  
        \caption{\carmonshortcit}
        \label{fig:carmon}
    \end{subfigure}
    \begin{subfigure}[b]{0.235\textwidth}
        \centering
        \includegraphics[width=\textwidth]{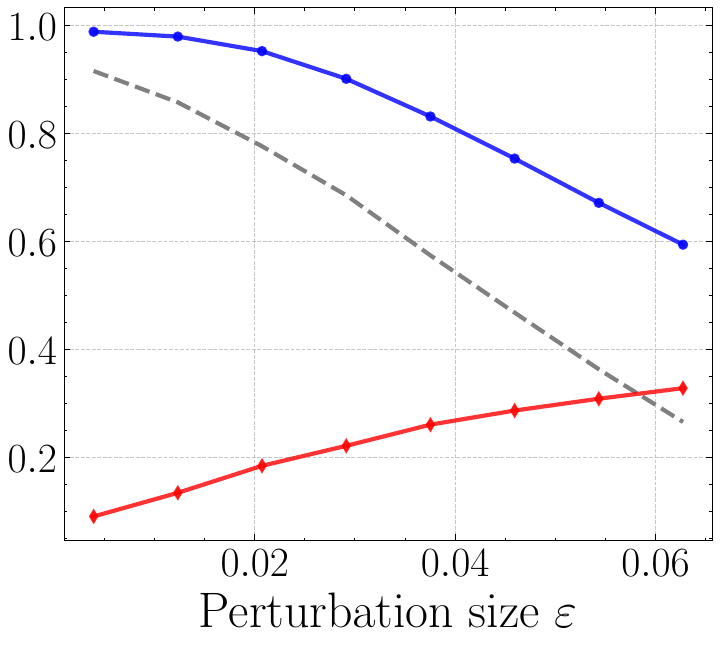}  
        \caption{\xushortcit}
        \label{fig:xu}
    \end{subfigure}
    \begin{subfigure}[b]{0.235\textwidth}
        \centering
        \includegraphics[width=\textwidth]{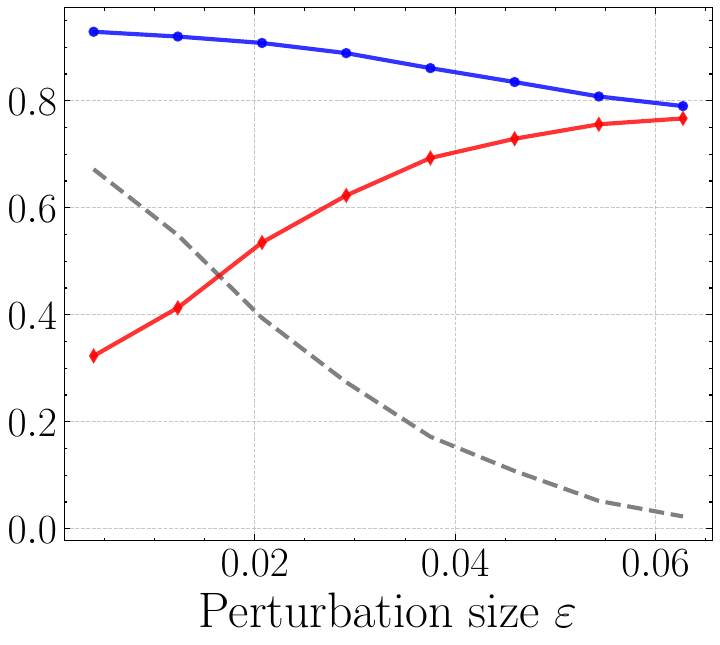}  
        \caption{\liushortcit}
        \label{fig:liu}
    \end{subfigure}
        \begin{subfigure}[b]{0.235\textwidth}
        \centering
        \includegraphics[width=\textwidth]{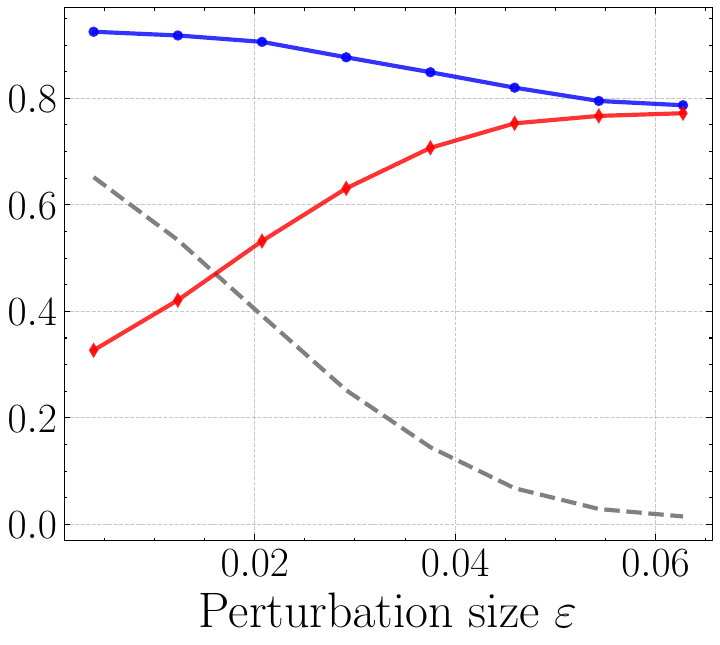}  
        \caption{\singhvitbshortcit}
        \label{fig:sing}
    \end{subfigure}
    \caption{Curves showing the baseline robust accuracy alongside \multishield's robust accuracy and rejection ratio under attacks with varying perturbation sizes ($\varepsilon$) across different robust models on CIFAR-10 (top two plots) and ImageNet (bottom two plots).}
    \label{fig:eps_plots}
\end{figure}

We present in \autoref{tab:multishield_effectiveness} the results on the effectiveness of \multishield in detecting adversarial examples across three distinct attack scenarios: (1) the baseline image classifiers is the target of attack without \multishield detection (Baseline); (2) \multishield is integrated during the final prediction while the image classifier is under attack (Multi-Shield); and (3) an adaptive attacker targets both the image classifier and \multishield, providing a worst-case evaluation of the defense mechanism (Multi-Shield - Adaptive Attack).
We then integrate our analysis with \autoref{fig:eps_plots} showing the performance of \multishield against adversaries of varying strengths $\varepsilon$.
Lastly, we compare \multishield with two state-of-the-art rejection defenses (i.e., \cite{melis2017deep} and \cite{sotgiu2020deep}) in \autoref{tab:comparison-soa}.\medskip
\begin{table*}[ht]
\centering
\caption{Comparison of \multishield, NR~\cite{melis2017deep}, and DNR~\cite{sotgiu2020deep} over five runs on 1,000-sample subsets from CIFAR-10. The rejection thresholds for NR and DNR are set to roughly match \multishield's accuracy. \emph{Robust Acc. (A)} refers to accuracy under an adaptive attack.\smallskip}
\label{tab:comparison-soa}
\setlength\tabcolsep{5.pt}
\begin{tabular}{@{}lclllclll@{}}
\toprule
Defense & Model & Clean acc. & Robust acc. & Robust acc. (A) & Model & Clean acc. & Robust acc. & Robust acc. (A) \\ 
\midrule
NR & \multirow{3}{*}{\standardshort} & 91.7 $\pm$ 0.002 & 3.2 $\pm$ 0.001 & 0.00 $\pm$ 0.00 &  \multirow{3}{*}{\carmonshort} & 88.4 $\pm$ 0.01 & 75.6 $\pm$ 0.03 & 59.2 $\pm$ 0.04 \\
DNR &  & 91.5 $\pm$ 0.001 & \textbf{96.1 $\pm$ 0.004} & 0.00 $\pm$ 0.00 &  & 88.2 $\pm$ 0.006 & 76.3 $\pm$ 0.02 & 60.6 $\pm$ 0.05 \\
Ours &  & 92.7 $\pm$ 0.002 & 91.9 $\pm$ 0.006 & \textbf{20.2 $\pm$ 0.003} &  & 88.3 $\pm$ 0.007 & \textbf{91.7 $\pm$ 0.009} & \textbf{64.2 $\pm$ 0.05} \\ 
\midrule
NR &  \multirow{3}{*}{\addepallishort} & 84.3 $\pm$ 0.02 & 72.1 $\pm$ 0.02 & 49.3 $\pm$ 0.05 & \multirow{3}{*}{\xushort} & 90.7 $\pm$ 0.007 & 75.9 $\pm$ 0.03 & 66.3 $\pm$ 0.04 \\
DNR &  & 84.2 $\pm$ 0.009 & 73.1 $\pm$ 0.04 & 50.8 $\pm$ 0.04 &  & 90.8 $\pm$ 0.009 & 76.0 $\pm$ 0.03 & 66.6 $\pm$ 0.03 \\
Ours &  & 83.8 $\pm$ 0.004 & \textbf{92.5 $\pm$ 0.008} & \textbf{58.9 $\pm$ 0.05} &  & 91.3 $\pm$ 0.002 & \textbf{88.8 $\pm$ 0.009} & \textbf{66.9 $\pm$ 0.07} \\ 
\midrule
NR & \multirow{2}{*}{\chenshort} & 82.6 $\pm$ 0.003 & 66.1 $\pm$ 0.01 & 43.8 $\pm$ 0.03 & \multirow{2}{*}{\gowalshort} & 82.3 $\pm$ 0.006 & 78.7 $\pm$ 0.03 & 62.1 $\pm$ 0.05 \\
Ours &  & 84.5 $\pm$ 0.003 & \textbf{91.9 $\pm$ 0.02} & \textbf{55.0 $\pm$ 0.03} &  & 87.0 $\pm$ 0.002 & \textbf{93.3 $\pm$ 0.008} & \textbf{69.6 $\pm$ 0.06} \\ 
\bottomrule
\end{tabular}
\end{table*}
\myparagraph{\multishield Effectiveness} 
We analyze the robustness of pre-trained models without \multishield (Baseline) and assess the impact of integrating \multishield to enable rejection of anomalous samples (Multi-Shield). The results in \autoref{tab:multishield_effectiveness} confirm that \multishield serves as an effective additional security layer significantly enhancing robust accuracy across all models without requiring their retraining. 
For example, \multishield raises the robust accuracy of non-robust models (i.e., \standardshort, \resnet, \wideres, \vit) from near $0\%$ (Baseline) to over $90\%$ across multiple datasets, effectively rejecting most adversarial examples.   
The robustness boost with \multishield is notable even on robust models, with an average improvement of $32\%$ on CIFAR-10 and $65\%$ on ImageNet (Multi-Shield).
The only trade-off is a slight reduction in clean accuracy, averaging approximately $1.8\%$ on CIFAR-10, $6\%$ on ImageNet and Caltech-101, $3\%$ on Food-101 and Oxford-IIIT Pets, and under $1\%$ on STL-10.\medskip
\myparagraph{Impact of Attack Strength on \multishield}
We evaluate \multishield against attackers of varying strengths (i.e., at different values of $\varepsilon$), demonstrating how robust accuracy and rejection ratio evolve as attack power increases. These trends are illustrated  in \autoref{fig:eps_plots} across four robust models (\carmonshort,\xushort,\liushort,\singhvitbshort) on CIFAR-10 (top two plots) and ImageNet (bottom two plots).
Firstly, in all three plots, the gap between the robust accuracy of \multishield (solid blue line) and the baseline (dashed gray line) widens as $\varepsilon$ increases, highlighting \multishield's superior robustness in response to increasing attack strength.
Secondly, the rejection ratio (red line) steadily increases in all three models, showcasing \multishield's ability to detect and reject stronger adversarial examples that mislead the baseline.
These results become more evident in the ImageNet models (\autoref{fig:liu}-\autoref{fig:sing}), where robust accuracy remains relatively more resilient while rejections rise significantly. \medskip
\myparagraph{Adaptive Attacker}
We conclude our analysis by evaluating \multishield under the worst-case scenario of an adaptive attacker who knows everything about the target model and exploits knowledge of the defense mechanism to bypass it~\citep{carlini2019evaluating}.
Specifically, we give the strong assumption of an adaptive attacker with complete knowledge of the rejection mechanism, including \multishield's image classifier, rejection mechanism, and inner workings (i.e., parameters and class prompts) to have a complete system evaluation.
Firstly, looking at the rightmost scenario in \autoref{tab:multishield_effectiveness} (Multi-Shield - Adaptive Attack), \multishield provides substantial improvements in robust accuracy across all non-robust models (\standardshort, \resnet, \wideres, \vit). 
While the gains are already significant for CIFAR-10 ($20.4\%$) and Oxford-IIIT Pets ($22.1\%$), they become even larger for Caltech-101 ($59.1\%$), Food-101 ($55.9\%$), and STL-10 ($43\%$).
For robust models, \multishield enhances robustness by an average of $3.3\%$ on CIFAR-10 and $5\%$ on ImageNet. This indicates that attacking \multishield, even under worst-case evaluation, requires significantly more effort and introduces further difficulties for the attacker. 
The constraint in \autoref{eq:adaptive_constraint} forces the attack to simultaneously deceive both the unimodal image classifier and the multimodal CLIP classifier, complicating the optimization process. We conclude that \multishield serves as an additional layer of defense to enhances the robustness of all considered models, both under non-adaptive and adaptive (worst-case) attack settings, across all datasets. \medskip
\myparagraph{Comparison with other Rejection Defenses}
We compare our method with state-of-the-art rejection defenses, including Neural Rejection (NR)~\citep{melis2017deep} and Deep Neural Rejection~\citep{sotgiu2020deep}, following the experimental setups outlined in their original works.
Experiments are conducted on five distinct subsets of $1000$ CIFAR-10 samples, reporting the average and standard deviation of results. 
The results, reported in \autoref{tab:comparison-soa}, showcase that \multishield consistently outperforms NR and DNR in robustness across various models. 
Notably, while DNR achieves high robustness under \standardshort (e.g., $96.1\%$), it fails entirely under adaptive attacks, dropping to $0.00\%$. 
In contrast, \multishield maintains significantly higher robustness against adaptive attackers (e.g., $20.2\%$ vs. $0.00\%$ for \standardshort).
Similar improvements are observed across other robust models, and the low standard deviation across multiple runs confirms the consistency of our results, reinforcing the validity of our conclusions. 
On \addepallishort, \multishield achieves $58.9\%$ robust accuracy under adaptive attacks, outperforming NR and DNR by approximately $8$ and $9$ percentage points, respectively. Likewise, on \gowalshort, \multishield reaches $69.6\%$ robust accuracy, surpassing NR ($62.1\%$) by $7.5\%$.
Beyond robustness, \multishield is also computationally less demanding. 
Both NR and DNR require training one or multiple one-vs-all SVMs for each class, leading to substantial memory and computational overhead. DNR, in particular, necessitates extreme batch size reductions (e.g., batch size = 1) to fit within 40GB of VRAM when applied to CIFAR-10, making it impractical for larger datasets (e.g., ImageNet) or more complex models (e.g., \chenshort, \gowalshort). 
We thus conclude that these comparisons further emphasize the benefits of \multishield as an additional security layer for robust classification systems, enhancing resilience against adversarial threats while maintaining high accuracy and computational efficiency.

\section{Conclusions, Limitations, and Future Works}\label{sec:conclusions}
In this work, we introduce \multishield, a novel defense that implements a rejection mechanism based on two core principles: (i) integration of multimodal information and (ii) interaction between adversarial training and a multimodal adversarial detector.
The first principle is realized by analyzing visual features and their semantic alignment with the textual prompts for the classes. 
In this way, \multishield enables the identification and rejection of adversarial examples when classification presents uncertainties or semantic inconsistencies. It leverages agreement in predictions between an image classifier and a multimodal model to decide whether to abstain from making a prediction.
Secondly, \multishield complements existing defenses that rely on adversarial training, providing an additional security layer against adversarial attacks. 
Our experiments show that \multishield consistently detects adversarial examples that mislead image classifiers and remains effective even under worst-case adaptive attacks.
%
Despite its strengths, \multishield may struggle to detect adversarial inconsistencies in datasets with abstract or meaningless labels, such as ``Class 1" or ``Class 2," as these provide no meaningful semantic cues for the multimodal model to align textual and visual data.
For future work, we aim to explore alternative methods for crafting more descriptive class label prompts, investigate their impact on robustness, and extend adversarial training to the multimodal model to further strengthen the defense.
\medskip
\myparagraph{Acknowledgments}
This work has been partially supported by the EU—NGEU National Sustainable Mobility Center (CN00000023), Italian Ministry of University and Research Decree n. 1033—17/06/2022 (Spoke 10), and project SERICS (PE00000014) under the MUR National Recovery and Resilience Plan funded by the European Union—NextGenerationEU. Lastly, this work was carried out while Dario Lazzaro was enrolled in the Italian National Doctorate on Artificial Intelligence run by the Sapienza University of Rome in collaboration with the University of Genoa.

\bibliographystyle{IEEEtranN}
\bibliography{main}
\end{document}